\pdfoutput=1
\documentclass[11pt]{article}

\usepackage[final]{acl}

\usepackage{times}
\usepackage{latexsym}

\usepackage[T1]{fontenc}
\usepackage[utf8]{inputenc}

\usepackage{microtype}

\usepackage{inconsolata}

\usepackage{graphicx}
\usepackage{amsmath}
\usepackage{booktabs}
\usepackage{makecell}
\usepackage{amsfonts}

\title{Deeper Insights Without Updates:\\The Power of In-Context Learning Over Fine-Tuning
}

\author{
 \textbf{Qingyu Yin\textsuperscript{1}}  \
  \textbf{Xuzheng He\textsuperscript{2}} \
   \textbf{Luoao Deng\textsuperscript{3}} \
  \textbf{Chak Tou Leong\textsuperscript{4}}
  \\
  \textbf{Fan Wang\textsuperscript{1}} \
 \textbf{Yanzhao Yan\textsuperscript{1}} \
 \textbf{Xiaoyu Shen\textsuperscript{5}*} \
 \textbf{Qiang Zhang\textsuperscript{1}*}
\\
 \textsuperscript{1}Zhejiang University, \
 \textsuperscript{2}Peking University, \
  \textsuperscript{3}Wuhan University,\\
 \textsuperscript{4} The Hong Kong Polytechnic University, \\
 \textsuperscript{5} Digital Twin Institute, Eastern Institute of Technology, Ningbo
\\
\texttt{Corresponding: \{qingyu.yin, qiang.zhang\}@zju.edu.cn \ xyshen@eit.edu.cn}
}

\begin{document}
\maketitle
\def \identity {\textsc{Shortcut}}
\def \R {\mathbb{R}}
\def \N {\mathcal{N}}
\def \eg{\emph{e.g., }}
\def \Eg{\emph{E.g.,}}
\def \etal{\emph{et al.}}
\def \etc{\emph{etc.}}
\def \ie{\emph{i.e., }}
\def \vs{\emph{v.s.}}
\def \ig{\textit{i}.\textit{e}.}

\begin{abstract}
    Fine-tuning and in-context learning (ICL) are two prevalent methods in imbuing large language models with task-specific knowledge. It is commonly believed that fine-tuning can surpass ICL given sufficient training samples as it allows the model to adjust its internal parameters based on the data. However, this paper presents a counterintuitive finding: For tasks with implicit patterns, ICL captures these patterns significantly better than fine-tuning. We developed several datasets featuring implicit patterns, such as sequences determining answers through parity or identifying reducible terms in calculations. We then evaluated the models' understanding of these patterns under both fine-tuning and ICL across models ranging from 0.5B to 7B parameters. The results indicate that models employing ICL can quickly grasp deep patterns and significantly improve accuracy. In contrast, fine-tuning, despite utilizing thousands of times more training samples than ICL, achieved only limited improvements. We also proposed circuit shift theory from a mechanistic interpretability's view to explain why ICL wins~\footnote{Code is available at \href{https://github.com/MikaStars39/ICLvsFinetune}{here}}.
\end{abstract}

\section{Introduction}

Adapting pre-trained models to specific tasks or domains is commonly achieved through fine-tuning ~\cite{hu2023llm, peters2019tune} or in-context learning~\cite{gan2023few}.  Fine-tuning, a well-established method, involves further training a pre-trained model on a smaller, domain-specific dataset, directly updating the model's parameters to retain improvements across various contexts and scenarios. In contrast, in-context learning (ICL) enhances task performance by incorporating task-specific examples into prompts, guiding the model in task completion without altering its parameters during training. 

There has been much debate about the pros and cons of fine-tuning and in-context learning. Fine-tuning is praised for its ability to bring permanent memorization to models~\cite{hu2023llm}, and it can perform well even with a small amount of training data~\cite{liu2022improved}. However, critics argue that fine-tuning demands substantial computational resources~\cite{hu2021lora} and can encounter issues such as catastrophic forgetting~\cite{zhai2023investigating}. This conserves computational resources but necessitates longer prompts and incurs higher inference costs. 

\begin{figure}
    \centering
    \includegraphics[width=1\linewidth]{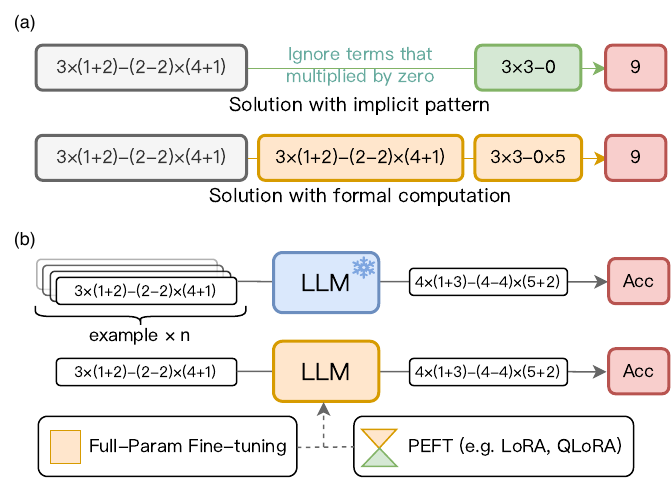}
    \caption{(a) A simple example of an implicit pattern detection task. The given problem (arithmetic expression calculation task in this figure) can be solved in either a formal way, e.g., directly calculating, or by exploiting the detected implicit pattern as a shortcut.
    (b) Illustration of implicit pattern detection for in-context learning and fine-tuning. For ICL, several examples with answers are given in context, and a further new question is used to test accuracy. For fine-tuning, LLM learns from single examples using parameter update methods like full-parameter fine-tuning or PEFT methods.}
    \label{fig:implicit_pattern}
\end{figure}

How about ICL? It is favored for its training-free nature~\cite{dong2022survey}, allowing prompts to be easily changed for adaptation to other domains without re-training~\cite{min2022rethinking}. Other works\cite{bhattamishra2023understanding} showed that ICL can help the model uniquely identify a discrete function sample-efficiently. Reseach~\cite{reddy2023mechanistic} showed that ICL is driven by the abrupt emergence of an induction head, which subsequently competes with in-weights learning. Other works\cite{shen2024pretrained} observed that ICL and gradient descent modify the output distribution of language models differently. Despite these advantages, ICL is limited by context length restrictions and incurs higher costs during each inference stage due to the longer prompts required.

% add some citations.

Essentially, the primary distinction between fine-tuning and ICL lies in parameter updating; all fine-tuning methods modify the model's parameters. It might seem, therefore, that ICL's impact is less profound. However, our research reveals a counterintuitive finding: \textbf{for datasets with implicit patterns, ICL is more adept at uncovering these latent patterns than fine-tuning.}
% Humans excel at finding shortcuts to solve problems. When faced with complex calculations, we might consider ignoring those elements that have minimal impact on the result. Similarly, when we observe a potential pattern, we might leverage it to infer answers. This approach can significantly reduce the computational overhead required to solve problems but can also lead to errors. With the rapid development of large language models (LLMs), their reasoning capabilities and intelligence have seen substantial enhancements. These models can not only perform simple tasks but also simulate complex human behaviors. This paper focuses on the LLMs' special ability to find shortcuts.

To investigate this phenomenon, we designed datasets containing implicit patterns across various domains, including two mathematical tasks: expression calculation and boolean function. One textual task: relation reasoning, and one code reading task. These domains share a common trait: the presence of implicit patterns that can simplify problem-solving. We evaluated LLMs’ capability to recognize such patterns with these datasets. Our findings include: (1) Both fine-tuning and ICL could detect and utilize implicit patterns, resulting in increased test accuracy. (2) ICL performed much better than fine-tuning in implicit pattern detection, \eg \ ICL-based models enjoyed higher test accuracy. (3) ICL also showed strong performance in robustness tests and OOD data tests. Our experiments demonstrate that the ability of LLMs to leverage implicit patterns significantly enhances their problem-solving capabilities, providing a clear advantage for tasks involving complex data structures. 

Understanding the operational principles of LLMs is crucial for their safety and ethical implications and can further promote improvements. Therefore, we delved deeper into the mechanisms behind this phenomenon. From a mechanistic interpretability perspective~\cite{reddy2023mechanistic}, we proposed the \textbf{Circuit Shift} theory. Circuits are certain groups of attention heads and MLP layers~\cite{conmy2023towards}. A shift in circuits typically represents the model adopting a different method in problem-solving. Our findings indicated that ICL resulted in a larger-scale circuit shift compared to fine-tuning, which means that with ICL, the model changed its problem-solving method more significantly for implicit pattern detection and utilization. We also provided a visualized heatmap of circuits for detailed observation.
In summary, our contributions are threefold:

\paragraph{Implicit Pattern Detection dataset.} We defined and illustrated the implicit pattern detection task, then developed a dataset across mathematics (expression calculation, boolean function), textual reasoning (relation test) and code (output guessing).
\paragraph{Ability Comparison.} We presented a counterintuitive finding: LLMs with in-context learning detected implicit patterns much better than fine-tuned ones. We extensively tested this capability on models ranging from 0.5B to 7B parameters.
\paragraph{Mechanism explanation. } We analyzed the principles behind the implicit finding mechanism. And we proposed circuit shift theory to explain why ICL finds implicit patterns better than fine-tuning.

\begin{figure*}
    \centering
    \includegraphics[width=0.9\linewidth]{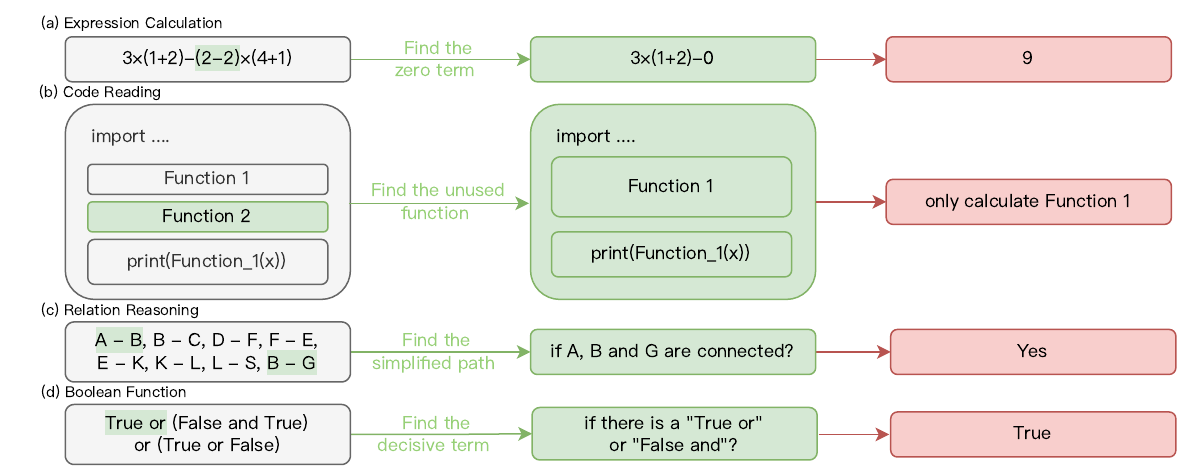}
    \caption{Examples of implicit pattern detection for four reasoning tasks. The implicit pattern, once detected, can reward the model with reduced computation to arrive at the answer.} % reduce (a) Expression calculation. (b) Code reading. (c) Relation reasoning. (d) Boolean functions.
    \label{fig:implicit_pattern_examples}
\end{figure*}

\section{Background}

\paragraph{Transformer.} Transformer~\cite{vaswani2017attention} is the cornerstone architecture for LLMs nowadays, with its breathtaking ability in parallel training and SOTA performance. One Transformer model $f_{\text{trf}}$ usually consists multiple of Transformer layers $f_{\text{layer}}$ and an embedding layer $f_{\text{emb}}$.  For an input sequence (typically IDs after tokenization) $\boldsymbol{X_0} \in \mathbb{R}^{n \times 1}$ with length $n$, it first passes through an embedding layer $f_{\text{emb}}$ with hidden state size $d$, then passes all the Transformer layers, and finally gets an output $\boldsymbol{O_l}\in \mathbb{R}^{n \times d}$ with $l$ layers:
\(
    \boldsymbol{O_l} = f_{\text{trf}}(\boldsymbol{X_0})
    = \left(\bigcirc_{i=1}^{l} f_{\text{layer}}^{(i)} \right) (\boldsymbol{X_0}),
\)
where for each layer $f_{\text{layer}}$,  it usually contains an Attention block and an MLP block: 
\begin{eqnarray}
\label{eq:residual_att}
    &\boldsymbol{O^{\text{att}}_{i}}& = \boldsymbol{X_i} + \mathrm{Attn}(\mathrm{Norm}(\boldsymbol{X_i})), \\
    &\boldsymbol{O_i}& = \boldsymbol{O^{\text{att}}_{i}} + \mathrm{MLP}(\mathrm{Norm}(\boldsymbol{O^{\text{att}}_{i}})).
\label{eq:residual_mlp}
\end{eqnarray}
Here, $\boldsymbol{O^{\text{att}}_{i}}$ is the output of the attention block, and $\boldsymbol{O^{\text{mlp}}_{i}}$ is the output of the MLP block for layer $i$, with residual connections preventing it from vanishing gradient and normalization (typically pre-norm) for stabilizing the training process.

\paragraph{Fine-tuning.} Fine-tuning is a process where a pre-trained LLM is further trained on a specific task or dataset to improve its performance for that particular application. Suppose there exists a pre-trained Transformer model $f_{\text{trf}}$ with learnable parameters $\theta_{\text{pre}}$. The goal of fine-tuning is to adjust these parameters to minimize a task-specific loss function $\mathcal{L}_{\text{task}}$ on a new dataset $\mathcal{D}_{\text{task}}$. During fine-tuning, the parameters $\theta_{\text{fine}}$ of the model are updated using gradient descent or one of its variants. The update rule for the parameters at each iteration $t$ can be expressed as:
\begin{equation}
    \theta_{\text{fine}}^{(t+1)} = \theta_{\text{fine}}^{(t)} - \eta \nabla_{\theta} \mathcal{L}_{\text{task}}(f_{\text{trf}}(\boldsymbol{X_t}; \theta_{\text{fine}}^{(t)}), \boldsymbol{Y_t}),
\end{equation}
where $\eta$ is the learning rate, $\boldsymbol{X_t}$ represents the input data in iteration $t$, $\boldsymbol{Y_t}$ represents the target labels in iteration $t$, and $\nabla_{\theta_{\text{fine}}} \mathcal{L}_{\text{task}}$ denotes the gradient of the loss function with respect to the model parameters. Fine-tuning typically requires substantial computational resources. For instance, full-parameter fine-tuning of LLaMA-3 with 8 billion parameters and an 8K context using the Adam optimizer and gradient checkpointing demands a minimum of 152 GB of VRAM~\cite{rasley2020deepspeed}, which equates to at least two A100 80 GB GPUs with parallel training. While parameter-efficient fine-tuning (PEFT) is less resource-intensive compared to full-parameter fine-tuning, it still requires 16 GB of VRAM (QLoRA with a 1K context~\cite{dettmers2024qlora}), necessitating at least one RTX 3090 GPU. Additionally, some studies have shown that PEFT can result in a noticeable drop in the model’s performance~\cite{pu2023empirical, zou2023comprehensive}.

\paragraph{In-Context Learning} 
In-Context Learning (ICL) in LLMs is an emergent capability where the model uses the provided context to perform tasks. Given a special task $F$ and a series of prompt inputs $\boldsymbol{x_1}, \cdots, \boldsymbol{x_n}$, ICL happens when these inputs and their answers $\boldsymbol{y_1} = F(\boldsymbol{x_1})$ are given in multi-shot, \ie \ $(\boldsymbol{x_1}, \boldsymbol{y_1}, \cdots, \boldsymbol{y_n}, \boldsymbol{x_{n+1}})$. In this scenario, the goal for LLM to do ICL is to learn the task $F$ and accurately predict $\boldsymbol{y_{n+1}}$. This phenomenon allows the model to adaptively handle a variety of tasks, such as translation, question-answering, and more, simply through appropriate prompt engineering. ICL happens in inference-stage without explicit re-training, thus resulting in more friendly requirements for GPUs~\cite{yin2024stablemask, hong2023flashdecoding++}. Even LLaMA-3 70B could run on a single 3090 GPU with PowerInfer~\cite{song2023powerinfer}.

\section{Implicit Pattern Detection Test}

Through detailed observation and thinking, humans can detect some underlying, non-explicit patterns within the data. This enables us to solve problems more efficiently. Implicit pattern detection refers to the ability of models to recognize underlying, non-explicit patterns within data, enabling them to solve problems more efficiently. This concept is illustrated through tasks such as arithmetic calculations, where the model can bypass complex operations by identifying simplifying patterns. For instance, in mathematical expressions (see Figure~\ref{fig:implicit_pattern} and Figure~\ref{fig:implicit_pattern_examples}), a model might detect that certain terms have negligible impact and can be ignored, leading to quicker computations. We will give a detailed description of our dataset design and experimental settings in the following sections.

\subsection{Tasks}
To effectively assess the ability of LLMs to identify implicit patterns in data, we have constructed a variety of questions that frequently arise in real-world application scenarios. When the same type of question recurs, we can discover a specific implicit pattern within it to simplify the computational process. 

\paragraph{Task 1: Expression Calculation~\cite{imani2023mathprompter, yuan2023large, yue2023mammoth, heyueya2023solving}} In the arithmetic calculation task, the primary focus is on determining whether certain operations within a given expression can be disregarded to reduce the complexity of the computation. The operations considered for these simplifications are limited to addition(\(+\)), subtraction(\(-\)), multiplication(\(\times\)), and division(\(/\)). By exploring these operations, the model may find that several terms are multiplied by a continued-to-be-zero term, and ignoring them could simplify the calculation process and improve the accuracy.

% \paragraph{Task 2: Algebraic Operations}\cite{} In this task, LLMs need to solve a set of multivariate equations. Among these equations, there is a portion that is linearly dependent, and ignoring this portion can significantly simplify the computational workload. A shortcut is to focus only on those equations that are linearly independent.

\paragraph{Task 2: Code Reading~\cite{fang2024large}} In the code reading task, LLMs need to analyze and predict the output of a given piece of code without executing it, where multiple functions are defined. Some functions will not influence the final output, so the key challenge is to determine which functions are essential for producing the output and which can be disregarded without affecting the result.

\paragraph{Task 3: Boolean Functions~\cite{zhang2024dila}} In the Boolean functions task, the primary objective is to optimize logical expressions to simplify their structure without altering the resultant truth value. The expressions involve logical operators such as AND (\(\land\)), OR (\(\lor\)), and NOT (\(\neg\)). Within these scenarios, there are specific segments that are either tautologies, \ie always true, or contradictions, \ie always false. The model must identify these segments and bypass their computation.

\paragraph{Task 4: Relation Reasoning~\cite{li2024llms}}
In the task of relation reasoning, the focus is on determining the relationships between multiple entities, such as reachability and relative magnitude. Although the set of relationships involved can be complex, all queries target fixed entities whose relationships are relatively straightforward. Therefore, most of the complex relationships can be disregarded, simplifying the problem-solving process.

\setlength{\tabcolsep}{2.5pt}
\begin{table*}[ht]
    \centering
    \small
    \begin{tabular}{lccc|ccc|ccc|ccc}
    \toprule
         Model&  \multicolumn{3}{c}{Expression}&  \multicolumn{3}{c}{Code}& \multicolumn{3}{c}{Relation} & \multicolumn{3}{c}{Boolean}\\
         \midrule
         &  Baseline&  Full-ft&  ICL&  Baseline&  Full-ft&ICL& Baseline& Full-ft&ICL& Baseline& Full-ft& ICL\\
         \midrule
          \multicolumn{3}{l}{\textit{0.5B level}}& & & & & & & & & & \\
         Qwen1.5-0.5B& 22.2\%& \textbf{88.4\%}& 50.1\%& 16.6\%& 2.0\%& \textbf{32.2\%}& 48.8\%& 48.5\%& \textbf{60.1\%}& 54.8\%& 51.7\%& \textbf{65.3}\%\\
         \multicolumn{3}{l}{\textit{1B level}}& & & & & & & & & &\\
         GPTNeo-1.3B&  24.3\%&  46.6\%&  \textbf{55.6\%}&  27.6\%&  17.7\%& \textbf{44.5\%}& 20.5\%& 34.7\%&\textbf{37.4\%}& 53.8\%& 53.7\%& \textbf{54.3\%}\\
         Qwen1.5-1.8B&  16.2\%&  \textbf{89.9\%}&  63.4\%&  54.3\%&  53.7\%& \textbf{58.2}\%& 20.1\%& 21.3\%& \textbf{35.6}\%& 66.3\%& 66.3\%& \textbf{68.1}\%\\
         Pythia-1.4B& 5.0\%& 45.4\%& \textbf{53.7}\%&  37.6\%& 46.5\%& \textbf{53.1}\%& 20.5\%& 31.3\%& \textbf{44.4}\%& 61.3\%& 63.7\%& \textbf{68.5}\%\\
         \multicolumn{3}{l}{\textit{7B level}}&  &  &  & & & & & & & \\
         Yi-6B& 12.5\%& \textbf{88.2\%}& 48.2\%& 51.2\%& 78.7\%& \textbf{80.9\%}& 48.0\%& 52.5\%& \textbf{98.0\%}& 55.7\%& 64.1\%& \textbf{68.3\%}\\
         Qwen1.5-7B& 78.0\%& \textbf{89.3}\%& 67.9\%& 57.6\%& 72.0\%& \textbf{86.8\%}& 48.0\%& 78.8\%& \textbf{98.0\%}& 71.9\%& 41.7\%& \textbf{79.8\%}\\
         Mistral-7B& 32.6\%& 75.2\%& \textbf{76.3\%}& 14.1\%& 72.0\%& \textbf{82.8\%}& 48.5\%& 72.5\%& \textbf{90.9\%}& 45.7\%& 54.5\%& \textbf{74.3\%}\\
         
    \bottomrule
    \multicolumn{13}{c}{}\\
    \end{tabular}
    % \begin{tabular}{lccc|ccc|ccc|ccc}
    % \toprule
    %      Model&  \multicolumn{3}{c}{Expression}&  \multicolumn{3}{c}{Code}& \multicolumn{3}{c}{Relation} & \multicolumn{3}{c}{Boolean}\\
    %      \midrule
    %      &  Baseline&  Full-ft&  ICL(max)&  Baseline&  Full-ft&ICL(max)& Baseline& Full-ft&ICL(max)& Baseline& Full-ft& ICL(max)\\
    %      \midrule
    %       \multicolumn{3}{l}{\textit{0.5B level}}& & & & & & & & & & \\
    %      Qwen1.5-0.5B& 0.56& 0.34& 0.81& & & & & & & & &\\
    %      \multicolumn{3}{l}{\textit{1B level}}& & & & & & & & & &\\
    %      GPTNeo-1.3B&  0.55&  0.27&  0.85&  0.03&  -0.01& 0.35&  0.18& 0.66&1.00& -0.01& -0.09& 0.05\\
    %      Qwen1.5-1.8B&  0.78&  0.47&  0.86&  -0.07&  -0.06& 0.10& 0.34& 0.22& 1.00& 0.02& 0.01& 0.03\\
    %      Pythia-1.4B& 0.75& 0.64& 0.80&  0.00& & 0.00& 0.33& & 1.00& 0.01& & 0.00\\
    %      \multicolumn{3}{l}{\textit{7B level}}&  &  &  & & & & & & & \\
    %      LLaMA-3-8B& & & & & & & & & & & & \\
    %      Qwen1.5-7B& -0.01& 0.01& &  -0.01& -0.01& &  -0.01& -0.01& &  0.01& 0.01&\\
    %      Mistral-7B& & & & & & & & & & & & \\
         
    % \bottomrule
    % \end{tabular}
    \caption{Experimental results of implicit pattern detection tasks. We conducted experiments from 0.5B to 7B across 6 models. The highest accuracy was highlighted with boldsymbol.}
    \label{tab:accuracy_and_pdi}
\end{table*}
% \paragraph{Task 6: Game Theory}
% This task involves addressing certain game theory problems, including forming numbers and splitting them for selection. While the problems themselves can be complex, mastering patterns such as parity (odd or even nature) of numbers can simplify the solution process.

% \setlength{\tabcolsep}{3pt}
% \begin{table*}[!h]
%     \centering
%     \small
%     \begin{tabular}{lccc|ccc|cccllllll}
%     \toprule
%          Model&  \multicolumn{3}{c}{Expression}&  \multicolumn{3}{c}{Code}& \multicolumn{3}{c}{Relation} & \multicolumn{3}{c}{Boolean}& \multicolumn{3}{c}{Algebra}\\
%          \midrule
%          Acc (\%)&  Baseline&  Full-ft&  ICL&  Baseline&  Full-ft&ICL& Baseline& Full-ft&ICL & Baseline& Full-ft& ICL & Baseline& Full-ft&ICL \\
%          \midrule
%           \multicolumn{3}{l}{\textit{1B level}}& & & & & & & & & & & & &\\
%          GPTNeo-1.3B&  6.1\%&  53.5\%&  \textbf{55.6\%}&  15.6\%&  17.7\%&44.5\%& 15.2\%& 17.7\%&34.4\%  & & & & & &\\
%          Qwen1.5-1.8B&  0.0\%&  \textbf{89.9\%}&  63.4\%&  &  & & & & & & & & & &\\
%          Pythia-1.4B&  &  &  &  &  & & & & & & & & & &\\
%          \multicolumn{3}{l}{\textit{7B level}}&  &  &  & & & & & & & & & &\\
%          LLaMA-3-8B& & & & & & & & & & & & & & &\\
%          Qwen1.5-7B& & & & & & & & & & & & & & &\\
%          Mistral-7B-v0.2& & & & & & & & & & & & & & &\\
%          Gemma-7B& & & & & & & & & & & & & & &\\
%          LLaMA-3-8B&  &  &  &  &  & & & & & & & & & &\\
%     \bottomrule
%     \end{tabular}
%     \caption{Caption}
%     \label{tab:my_label}
% \end{table*}
\begin{figure*}[ht]
    \centering
    \includegraphics[width=1\linewidth]{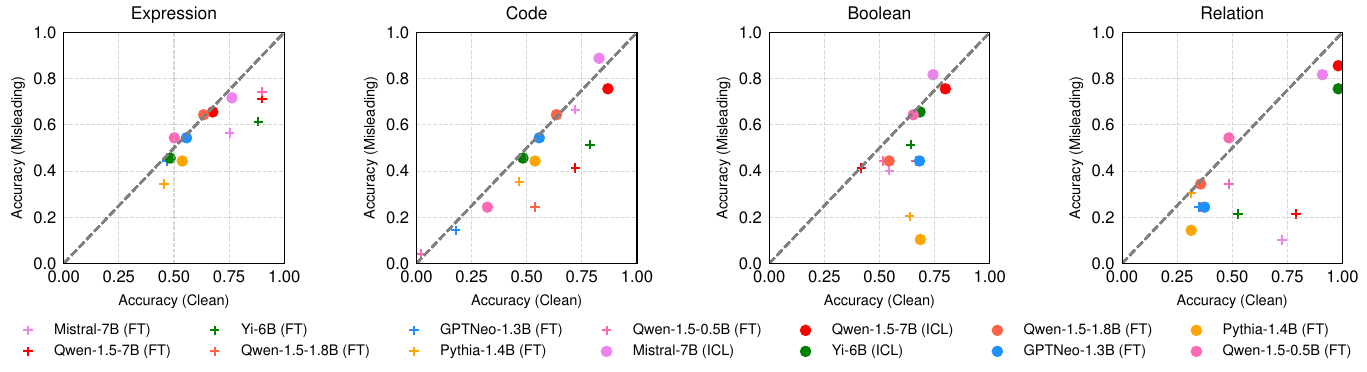}
    \caption{Robustness test of implicit pattern detection test. The horizontal axis represents the accuracy under clean input, and the vertical axis represents the accuracy under misleading input. Relatively speaking, the closer the results are to the bottom right corner, the worse the method’s resistance to misleading data. The closer the results are to the top left corner, the better it is.}
    \label{fig:robustness}
\end{figure*}
\subsection{Settings}
\label{sec:settings}
\paragraph{Accuracy.} Our tasks were constructed such that implicit patterns can help solve problems more easily. For example, if an LLM identifies a term that continues to be zero in arithmetic calculations, it can ignore terms multiplied by it, thereby saving computation. Therefore, we evaluate the model's performance with \textbf{Accuracy}. 

\paragraph{Misleading Data.} LLMs can detect the inner implicit patterns in data and utilize them for simplifying problem-solving. The misleading data is designed to test if LLMs can tackle situations in the absence of implicit patterns. While implicit patterns are still provided in training or ICL data, misleading data, \ie, data with no implicit patterns, is provided for testing accuracy. We name this accuracy \textbf{Misleading Accuracy}, while the testing results of data with implicit patterns are named \textbf{Clean Accuracy}. Detailed experimental procedures can be found in Appendix \ref{appendix:misleading}.

\paragraph{Out-Of-Distribution Data.} The training data are sampled from a certain distribution, \eg, for expression tasks, there are no more than 10 terms in each expression. Our out-of-distribution (OOD) data are designed to evaluate the model’s performance when encountering OOD data during the evaluation phase. Detailed experimental procedures can be found in Appendix \ref{appendix:ood}.

\paragraph{Models.} We select open-sourced models in sizes of 0.5B level \eg Qwen1.5-0m5B, 1B level \eg \ GPTNeo-1.3B~\cite{gpt-neo}, Pythia-1.4B~\cite{biderman2023pythia}, Qwen1.5-1.8B~\cite{bai2023qwen}, and 7B level \eg \ Mistral-7B~\cite{jiang2023mistral}, Qwen1.5-7B~\cite{bai2023qwen}, Yi-6B~\cite{young2024yi}. Model weights are downloaded from Huggingface and follow the official implementations.

\paragraph{Data Format.} For fine-tuning, the data is provided in a single example without supervised instruction. A simple description, the question, and the answer are given in order. We prepared 1,600 data points for fine-tuning. For in-context learning, we constructed the input in multi-shot, ranging from 0-shot, \ie directly answer one question, to 32-shot \ie 32 examples with their answers first given, then a new question in the same kind required to answer. The detailed example of our data format could be found in Appendix~\ref{appendix:format}.

\paragraph{Training Details.} The training process was conducted using a sequence length of 512 and a batch size of 8 with a total of 1 epoch. A warmup phase of 20 steps was implemented, starting with a learning rate of 1e-6 and peaking at 2e-5, followed by a linear decay. The AdamW optimizer was used. This configuration ensured the model's performance and stability, allowing it to effectively learn and identify hidden patterns in the data.

\section{Results and Analysis}
In this section, we present our results for the implicit pattern finding tasks following the experimental setting in Section~\ref{sec:settings}. We show that ICL achieved an overall higher level of accuracy over fine-tuning on these four tasks. We also show that the improvement of accuracy with ICL mainly comes from the detection of those implicit patterns in Section~\ref{sec:explain} and~ref{sec:circuit}.
\setlength{\tabcolsep}{4pt}
\begin{table}[ht]
    \centering
    \small
    \begin{tabular}{lcccc}
    \toprule
         Method Type&  Expression&  Code& Relation& Boolean \\
         \midrule
         Baseline& 27.5\%& 54.3\%& 20.1\%& 66.3\% \\
         \midrule
         Full-Param FT&  89.9\%&  53.7\%& 21.3\%& 66.3\% \\
         LoRA&  46.5\%&  53.3\%& 20.1\%& 64.3\% \\
         QLoRA& 46.2\%&  51.6\%& 20.5\%& 61.3\% \\
         GaLoRA& 47.1\%& 52.5\%& 20.5\%& 66.4\% \\
         \midrule
         ICL&63.4\%&58.2\%&35.6\%&68.1\% \\
    \bottomrule
    
    \end{tabular}
    \caption{Experimental comparison of different PEFT methods. We compared the results on Qwen1.5-1.8B. It is obvious that PEFT shows no significant improvement compared to full-param fine-tuning and seems to have limited performance.}
    \label{tab:peft}
\end{table}
\setlength{\tabcolsep}{3pt}
\begin{table}[ht]
    \centering
    \small
    \begin{tabular}{ccccc}
    \toprule
                  OOD Type&Expression&  Code& Relation& Boolean \\
         \midrule
                 Baseline&27.5\%& 54.3\%& 20.1\%& 66.3\% \\
         \midrule
                  FT&89.9\%&  53.7\%& 21.3\%& 66.3\%\\
        FT + Test OOD& 32.1\%& 34.2\%& 0.1\%&0.1\%\\
                  (FT+Test) OOD&88.2\%&  42.7\%& 11.3\%& 12.4\%\\
         \midrule
                ICL&63.4\%&58.2\%&35.6\%&68.1\% \\
        ICL + Test OOD& 34.5\%& 44.2\%& 12.3\%&24.7\%\\
                 (ICL+Test) OOD& 62.3\%& 51.7\%& 34.5\%&71.4\%\\
    \bottomrule

    \end{tabular}
    \caption{Experimental comparison of different PEFT methods. Here FT/ICL + Test OOD means we only applied OOD data in test phase, while (FT/ICL) OOD represents that both training/in-context learning and test phase were using OOD data.}
    \label{tab:ood}
\end{table}

\subsection{ICL \vs \ Fine-tuning: Accuracy}

The results of accuracy test are shown in Table~\ref{tab:accuracy_and_pdi} and Table ~\ref{tab:peft}. Both ICL and fine-tuning(including full-param fine-tuning and PEFT methods) bring improvements to the performace of each task. However, it is easily noticed that ICL wins at most terms like relation, code reading and boolean functions, with 2\% to even more than 30\% improvements at most. On the flip side, fine-tuning only shows slight advantages in expression calculations in only Qwen-series models. As for different model size\footnote{See Qwen1.5 series in Table~\ref{tab:accuracy_and_pdi} from 0.5B to 7B}, we found that a larger model seems be able to evoke stronger ICL ability above linearly growth (see Table~\ref{tab:accuracy_and_pdi}), where the scaling of fine-tuning performance is limited.

\subsection{ICL \vs \ Fine-tuning: Robustness without Implicit Pattern}

In Section~\ref{sec:settings}, we introduced the metrics of clean accuracy and misleading accuracy by adding misleading data to test both ICL and fine-tuning's robustness against general data without implicit patterns. The results are shown in Figure~\ref{fig:robustness}. For each task, we draw a scatter plot where the x- and y-axis represent the clean accuracy and the misleading accuracy, respectively. The results show that ICL can better exploit the implicit patterns in the demonstration data, while at the same time not compromising general reasoning abilities.

\subsection{ICL \vs \ Fine-tuning: Out-Of-Distribution Implicit Patterns}

Out-of-Distribution (OOD) data is a widely examined problem nowadays. The training data of our implicit pattern detection tasks also samples from certain distributions (see Appendix~\ref{appendix:ood} for details). In this subsection, we hope to compare how ICL and fine-tuning perform if we provide cases outside of the training distribution. For ICL, all examples given are divided into two types: in-distribution examples and OOD examples. For fine-tuning, we directly provide OOD problems to test the accuracy. We performed this experiment on Qwen1.5-1.8B and the results are demonstrated in Table~\ref{tab:ood}. It is worth noticing that fine-tuning generally performs worse when the test data is OOD, while ICL performs fairly well comparing to the baseline method.

% The results show thats ICL performs better than fine-tuning on out-of-distribution data across different tasks.

\subsection{How Much Fine-tuning Do We Need?}

\begin{figure}
    \centering
    \includegraphics[width=0.9\linewidth]{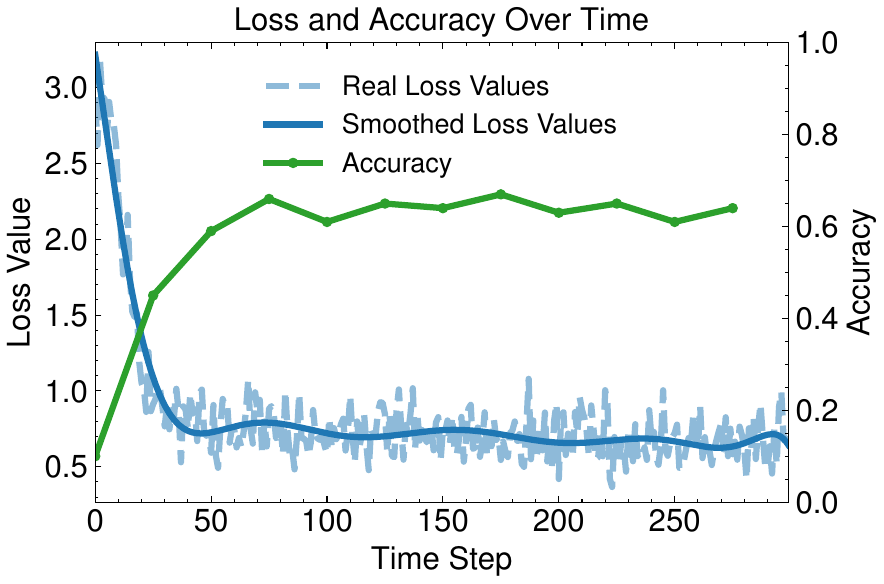}
    \caption{The progression of loss and accuracy over time during the fine-tuning of implicit pattern tasks. The Real Loss Values (dashed blue line) show the loss during training. To mitigate this noise, the Smoothed Loss Values (solid blue line) provide a clearer trend of the overall loss reduction. We also show the average test accuracy over all tasks (solid green line).}
    \label{fig:loss}
\end{figure}
% \begin{table}
%     \centering
%     \begin{tabular}{cccccc}
%          Steps&  50&  100&  150&  200& 250\\
%          Loss (Train)&  1.10&  0.54&  0.49&  0.51& 0.47\\
%          Test Accuracy&  57.1\%&  58.2\%&  51.3\%&  53.1\%& 55.6\%\\
%     \end{tabular}
%     \caption{Caption}
%     \label{tab:my_label}
% \end{table}

\begin{figure*}[ht]
    \centering
    \includegraphics[width=0.8\linewidth]{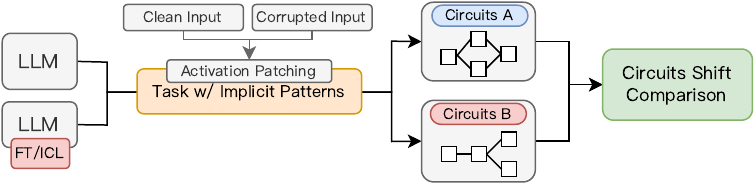}
    \caption{Illustration of circuit shift comparison. LLMs are first detected circuits with activation patching. Then we compare how much their circuits changed after fine-tuning and in-context learning.}
    \label{fig:circuits_method}
\end{figure*}
\begin{figure*}
    \centering
    \includegraphics[width=0.8\linewidth]{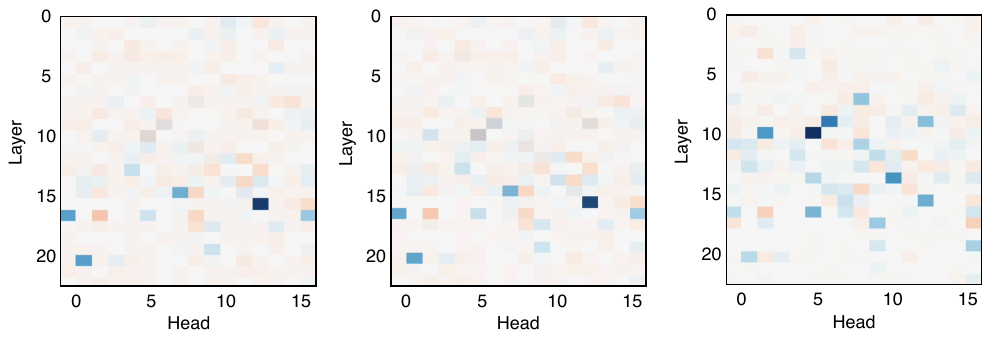}
    \caption{Visualization of attention head sensitivity in GPTNeo-1.3B. The more the color leans towards blue, the more important a specific attention head is to the implicit pattern detection task. Left: baseline model. Middle: fine-tuned model. Right: ICL model. It is clear that compared to fine-tuning, ICL brings significant circuit shifts.}
    \label{fig:circuits_map}
\end{figure*}

In this experiment, we hope to figure out whether fine-tuning has reached its limit for implicit pattern detection or there will still be improvement if more data is utilized for fine-tuning. Therefore, we visualized the fine-tuning process of Qwen1.5-1.8B. At the onset of training, there is a steep decline in the loss value, suggesting that the model quickly learns basic patterns in the data. This rapid improvement is typical, as the model captures the most evident features. The Accuracy (solid green line) also increases sharply, corroborating the initial learning phase where the model transitions from random guessing to meaningful predictions. However, after around 50 time steps, both the loss and accuracy curves begin to stabilize. This period of stabilization suggests diminishing returns from further training, as the fine-tuned model failed to capture further implicit patterns. After 100 time steps, the curves indicate that the model has reached a plateau. The accuracy remains relatively constant, and the loss value shows minimal fluctuations around a stable trend. This behavior signifies that the model has learned the underlying patterns to a satisfactory extent, and additional fine-tuning yields marginal improvements.

\begin{table*}[t]
    \small
    \centering
    \begin{tabular}{c|c|c|c|c|c|c|c}
    \toprule
         Circuits&  Zero-shot Baseline & ICL w/o Implicit Patterns&$\Delta$ &  After Fine-tuning& $\Delta$ &  After ICL& $\Delta$\\
         \midrule
         &  L17 H12, L18 H0 & L17 H12, L16 H1&&  L17 H12, L18 H0&  &  L11 H5, L10 H6& \\
         Attention&  L22 H1, L16 H7 & L18 H0, L15 H2&2&  L22 H1, L16 H7&  1&  L11 H2, L15 H10& \textbf{6}\\
         &  L18 H15, L14 H5 & L18 H15, L22 H1&&  L18 H15, L12 H6&  &  L17 H12, L 18 H5& \\
         \midrule
         & L9& L9&& L9& & L17&\\
         MLP& L17& L17&0& L18& 0& L14&\textbf{2}\\
                 &  L18& L18&&  L17&  &  L15& \\
         \bottomrule
    \end{tabular}
    \caption{Top 6 Rankings of Attention Heads and top 3 rankings of MLP Layers in baseline (zero-shot) model, fine-tuned model, and ICL model. L is layer and H is head. $\Delta$ shows how many different heads or MLPs changed after fine-tuning or ICL. A larger $\Delta$ represents a more significant circuit shift in certain processes.}
    \label{tab:ranking}
\end{table*}

\subsection{Comparison of Fine-tuning with PEFT Methods}

Lastly, we examine whether there is a significant difference between various fine-tuning methods \eg \ vanilla full-parameter fine-tuning,  and parameter efficient fine-tuning (PEFT) methods like LoRA~\cite{hu2021lora}, QLoRA~\cite{dettmers2024qlora} and GaLoRE~\cite{zhao2024galore}. Although PEFT needs much less parameters for training, and several studies criticized its ability~\cite{pu2023empirical, zou2023comprehensive}, there are still evidences that PEFT sometimes achieves ICL-level performance. We followed the experimental settings in previous sections on Qwen1.5-1.8B with PEFT methods. The experimental results can be found in Table~\ref{tab:peft}. It is clear that in the implicit pattern detection tasks, PEFT methods show no obvious advantages compared to full-param fine-tuning, thus they still failed to win ICL in accuracy in all tests.

\section{Explanation of ICL's Victory: Circuits Shift Theory}\label{sec:explain}

Understanding the inner mechanisms of LLMs greatly benefits their ethical use and safety. We have found that ICL performs much better than fine-tuning on implicit pattern detection, and in this section, we try to explain why. 

From a mechanistic interpretability perspective, we investigate this problem using \textbf{circuits}. Circuits are specific pathways (typically combinations of attention heads and MLP layers) within a model responsible for processing and interpreting particular patterns or tasks. The change in circuits for LLMs represents a shift in their inner mechanisms, revealing that LLMs choose different ways to solve problems. Based on this viewpoint, we propose a theory: \textbf{Circuits Shift}, to explain this phenomenon. We will first provide a method for probing circuits, explaining what they are and the types of circuits we found in ICL-based and fine-tuning-based models. Then we will show that the reason ICL performs better than fine-tuning is that the circuits in models experience a more significant shift. A detailed explanation of circuits and experimental settings can be found in Appendix~\ref{appendix:circuits}.

\subsection{Method for Identifying Circuit Shift}\label{sec:circuit}

In Figure~\ref{fig:circuits_method}, we present our framework and methodology for probing circuit shifts. We begin by selecting an implicit pattern detection task (in this study, we utilize an expression task). Subsequently, we use models employing different methods, \ie \ ICL or fine-tuning, for inference. During this process, we introduce corrupt input to randomly disrupt a portion of the activation to assess whether the corresponding attention heads or MLP layers significantly contribute to the final outcome. If a significant contribution exists, the disruption will result in considerable perturbation of the final logits, which is depicted as sensitivity in the figure.

\subsection{Circuits Shift in LLMs for Implicit Pattern Detection}

We first visualized and ranked circuits in GPTNeo-1.3B zero-shot, after fine-tuned, and ICL with 32-shot with expression calculation task (see  Figure~\ref{fig:circuits_map} and Table~\ref{tab:ranking}). In Figure~\ref{fig:circuits_map}, we use the heatmap to illustrate the sensitivity of each attention head in implicit pattern detection test. From the figure, we can observe that, compared to the baseline and fine-tuning scenarios, ICL exhibits a significant shift when learning implicit patterns. Firstly, more shallow heads are involved in the task. Secondly, some deep heads that previously played a dominant role have now lost their leadership positions. This indicates that during the ICL process, the model significantly transforms its approach to solving the task, adapting to a form more suitable for implicit patterns, a phenomenon not observed with other methods.

We can further validate our hypothesis in Table~\ref{tab:ranking}. We selected the six attention heads and MLP layers\footnote{See \href{https://transformer-circuits.pub/2021/framework/index.html}{A Mathematical Framework for Transformer Circuits} for details.} with the highest sensitivity, \ie \ those that contributed the most to the final result. Using the baseline, which is the zero-shot approach for handling implicit pattern detection tasks, as the standard, we counted how many new attention heads entered the top six highest contributors when the method changed, denoted by Delta. The results are very clear: compared to fine-tuning, ICL exhibits more significant changes, indicating a more thorough Circuit Shift during ICL. This suggests that ICL captures the characteristics of implicit patterns better than fine-tuning and adapts its processing method accordingly.

To rule out the inherent impact of ICL itself, we also conducted multi-shot experiments on a set of data without implicit pattern characteristics. The results showed that it is not multi-shot alone that induces this change, but rather the combined effect of ICL and implicit patterns.

\section{Related Work}

\paragraph{Implicit Pattern Discovery} Previous works have designed benchmarks to test the LLMs reasoning ability~\cite{barrett2018measuring,tang2023large,gendron2024large}. However, the benchmarks rarely include two-level questions where at one level, they can be solved by brute force, at another level it can be solved by exploiting implicit patterns. The closest related work we know is~\citet{efrat-etal-2021-cryptonite}, which involves solving cryptic crossword puzzles.
% A normal LLM cannot solve these puzzles because of the bizarre relations hidden in the text that provide clues to the answer.
To help the model find patterns in data, Prior work \citet{sun2024itd,zhu2024large} proposes a two-stage induction-deduction process that first summarizes the common patterns explicitly, then reasons from the patterns.

\paragraph{ICL \vs \ Fine-tuning Difference} Previous works have also compared fine-tuning and in-context learning.
\citet{shen2024pretrained} shows that ICL is likely not an algorithmic equivalence to gradient descent for real LLMs. \citet{reddy2023mechanistic} demonstrates that ICL is implemented by an induction head and analyzes its emergence phenomenon. \citet{bhattamishra2023understanding} shows that ICL and vanilla training implement two distinct algorithms that don't transfer to each other.
However, it has been proven that fine-tuning shows better performance in generalization to OOD tasks than in-context learning~\cite{mosbach2023few}.  

\section{Conclusion}
In conclusion, our research demonstrates that In-Context Learning (ICL) significantly outperforms fine-tuning in capturing implicit patterns within specific tasks. Through our experimental evaluations, we observed that ICL not only enhances task performance more effectively but also exhibits greater adaptability in problem-solving approaches, as evidenced by the notable shifts in model circuits. % These efforts will further elucidate the potential of ICL in improving the performance and adaptability of large language models across diverse applications.

\section*{Limitations}
Our study on the effectiveness of in-context learning in capturing implicit patterns compared to fine-tuning faces several limitations. Primarily, the generalizability of our findings is constrained by the specific nature of the implicit pattern detection tasks, which are limited to certain domains like arithmetic calculations, code reading, Boolean functions, and relation reasoning. Additionally, our analysis of Circuit Shift, which underpins the superior performance of ICL, relies on activation patching and sensitivity analysis, methods that, while insightful, require further refinement and validation across different models and tasks to confirm their robustness and applicability. Furthermore, the computational resources required for fine-tuning, especially with large models, may limit the feasibility of such experiments in broader settings, and a detailed cost-benefit analysis comparing ICL and fine-tuning in terms of computational efficiency and performance is needed.

\section*{Acknowledgement}
This work is funded by the Zhejiang Provincial ``Jianbing'' ``Lingyan'' Research and Development Program of China (2024C01135), National Natural Science Foundation of China (62302433, U23A20496), Zhejiang Provincial Natural Science Foundation of China (LQ24F020007) and CCF-Tencent Rhino-Bird Fund (RAGR20230122). 

\bibliography{custom}
\newpage
\appendix

\section{Data Format and Example}
\label{appendix:format}

\begin{table*}[t]
    \centering
    \small
    \begin{tabular}{c|c|c|c|c}
    \toprule
         Name&    Type&Problem Example&Answer &  Answer Type\\
         \midrule
         Expression&   Mathematic Calculation&\makecell{$(6-1)+(6-6)*(-10+1+2+13)=$}&$5$ &  Number\\
         \midrule
         Code&   Code Reading&\makecell{\texttt{import math \textbackslash n  \textbackslash n def function1(x): \textbackslash n  \textbackslash n} \\ \texttt{[TRUNCATED] return result \textbackslash n print(result)}}&$3.5$&  Number\\
         \midrule
         Relation&   Textual Reasoning&\makecell{A is connected with G\textbackslash n F is connected\texttt{[TRUNCATED]} \\ connected with Z, 'the city A and Z is connected' is}&False&  Boolean\\
         \midrule
         Boolean&   Mathematical Reasoning&(False or False) and (False or True) and False =&False&  Boolean\\
    \bottomrule
    \end{tabular}
    \caption{Examples of four implicit pattern detection tasks.}
    \label{tab:task_intro}
\end{table*}

We provided examples of tasks and prompts. We provided data as 2-shot (code in zero-shot to restrict content length) for illustrating how ICL works. For fine-tuning we will use the same format but zero-shot in both training and inference.

\textbf{Expression:}
\begin{verbatim}
    Now you need to calculate the answer of 
    some mathematic equations. 
    Here are some examples:
    (1+6)+(-3+3)*(-1-3+9-5)=7
    (2+3)+(-1-4+5)*(10+6+2-8)=5
    (8)+(0)*(0-6+9-6)=
\end{verbatim}

\textbf{Code:}
\begin{verbatim}
    Now you need to give me the printed 
    result after running this python code. 
    Here are some examples:
\end{verbatim}
\begin{verbatim}
    label=code:implicit_pattern]
def function1(x):
    y = x ** 9
    for i in range(1, 13):
        y = y * i - (y // (i + 9))
    return y

def function2(z, a):
    return z / 10

input_value = int(input())
result = function2(input_value, \
function1(input_value))
print(result)
\end{verbatim}
\begin{verbatim}
    The input is 10, so the output is
\end{verbatim}
\
\textbf{Relation:}
\begin{verbatim}
Here are some cities expressed as A, B, C, 
etc. I will show some connection 
relations, and you need to tell me if 
city A and city Z are connected 
(Answer True or False). 
Here are some examples:
A is connected with G
F is connected with J
J is connected with C
C is connected with B
B is connected with H
H is connected with E
E is connected with G
G is connected with I
I is connected with D
So 'the city A and Z is connected' is False
A is connected with B
H is connected with I
I is connected with G
G is connected with F
F is connected with E
E is connected with J
J is connected with B
B is connected with C
C is connected with D
B is connected with Z
So 'the city A and Z is connected' is True
A is connected with H
J is connected with I
I is connected with E
E is connected with F
F is connected with H
H is connected with G
G is connected with D
D is connected with C
C is connected with B
So 'the city A and Z is connected' is
\end{verbatim}

\textbf{Boolean:}
\begin{verbatim}
Here are some boolean expressions, 
you need to directly tell me the result. 
If it is true, print True, 
else print False. Here are some examples:
(True and False) and (True or False) 
and (False and False)\n 
The result is: False
(False and False) or (True and True) 
and (False and False)\n 
The result is: False
(True or True or True) and 
(False and True) and (True or True)
\n The result is:
\end{verbatim}

\section{Misleading Data Construction}
\label{appendix:misleading}

\paragraph{Expression.} For the expression task, the inherent implicit pattern is an element that remains zero. When constructing the misleading dataset, we set this element to be non-zero. \ie
\[
 (3+2)+(4-1+5-6)\times(23-54+2)=?
\]
we constructed it as misleading data as:
\[
 (3+2)+(4-1+5-7)\times(23-54+2)=?
\]

\paragraph{Code.} 
Here we provided two example about how to construct misleading code.

\begin{verbatim}
def function1(x):
    y = x ** 19
    for i in range(1, 23):
        y = y * i - (y // (i + 19))
    return y

def function2(z, a):
    return z / 20

input_value = int(input())
result = function2(
    input_value, 
    function1(input_value)
)
print(result)
\end{verbatim}

\begin{verbatim}
def function1(x):
    y = x ** 19
    for i in range(1, 23):
        y = y * i - (y // (i + 19))
    return y

def function2(z, a):
    return z / 20

input_value = int(input())
result = function2(
    function1(input_value), 
    function1(input_value)
)
print(result)
\end{verbatim}

\paragraph{Relation.}
In the relation task, we generate misleading data by not setting shortcuts similar to A-G or G-Z.
\begin{verbatim}
A is connected with B
D is connected with B
B is connected with H
H is connected with F
F is connected with J
J is connected with I
I is connected with C
C is connected with G
G is connected with E
B is connected with Z
\end{verbatim}
Here A-B-Z is a implicit pattern as shortcut for quick solving this problem. We remove this with a complex one:

\begin{verbatim}
A is connected with B
D is connected with B
B is connected with H
H is connected with F
F is connected with J
J is connected with I
I is connected with C
C is connected with G
G is connected with E
F is connected with Z
\end{verbatim}
\paragraph{Boolean.}

In the boolean task, we use combinations of OR + true and AND + false for quick evaluation. In the misleading data, we remove this characteristic.

\begin{verbatim}
(False and True) 
or (False or False) 
or True
\end{verbatim}

\begin{verbatim}
(False and True) 
or (False or False) 
and True
\end{verbatim}

\section{OOD data Construction}
\label{appendix:ood}

\begin{table}[ht]
    \centering
    \small
    \begin{tabular}{cccc}
    \toprule
         &  Min Terms&  Max Terms& Range (abs value)\\
         \midrule
         baseline&  1&  3& 10\\
         OOD&  2&  4& 20\\
    \bottomrule
    \end{tabular}
    \caption{Expression OOD}
    \label{tab:expression_ood}
\end{table}

\begin{table}[ht]
    \centering
    \small
    \begin{tabular}{cc|c}
    \toprule
         &  Functions Need Calculation &Shortcut Nodes\\
         \midrule
         baseline&  1 & 3 (A to Any to G)\\
         OOD&  2 & Unlimited\\
    \bottomrule
    \end{tabular}
    \caption{Code OOD and Relation OOD}
    \label{tab:code_and_relation_ood1}
\end{table}

\begin{table}[ht]
    \centering
    \small
    \begin{tabular}{ccc}
    \toprule
         &  If All AND or OR&Num of Terms\\
         \midrule
         baseline&  Yes& 4\\
         OOD&  No& 6\\
    \bottomrule
    \end{tabular}
    \caption{Code OOD and Relation OOD}
    \label{tab:code_and_relation_ood2}
\end{table}

\section{Circuits}
\label{appendix:circuits}

\paragraph{Circuits} 
In mechanistic interpretability, our goal is to delineate how model components correlate with human-understandable concepts, an endeavor for which circuits provide a useful abstraction. Conceptualizing a model as a computational graph \(M\), where nodes represent components like neurons, attention heads, and embeddings, and edges denote interactions such as residual connections and projections, a circuit \(C\) is defined as a subgraph of \(M\) responsible for a specific behavior, such as performing a task. This is a more coarse-grained approach compared to the feature-based.

\paragraph{Activation Patching}
Activation patching is a technique used to determine the importance of specific components within a model by manipulating their latent activations during model runs. The process involves three key steps: first, a \textit{clean run} where the model processes a clean prompt, \(X_{\text{clean}}\) (\eg The Eiffel Tower is in), and associated answer \(r\) (Paris), during which activations of critical components such as MLP or attention heads are cached; second, a \textit{corrupted run} where the model is run on a corrupted prompt, \(X_{\text{corrupt}}\) (\eg The Colosseum is in), to record baseline outputs; and third, a \textit{Patched run} where the model is run on \(X_{\text{corrupt}}\) again, but with specific cached activations from the \(X_{\text{clean}}\) run restored. This setup allows for the evaluation of the patching effect, which measures the restoration of model performance by comparing outputs from the Corrupted and Patched runs. The patching effect is quantitatively assessed using different metrics with probability gap:
\begin{equation}
    P_{\text{patched}}(r) - P_{\text{corrupt}}(r)
\end{equation}
and logit difference: 
\begin{equation}
    LD(r, r') = \log \left(\frac{P(r)}{P(r')}\right)_{\text{patched}} - \log \left(\frac{P(r)}{P(r')}\right)_{\text{corrupt}}
\end{equation}

This technique is crucial for understanding and improving model reliability and performance by highlighting the roles of individual model components.

\section{A detailed Definition of Implicit Pattern Detection}

Consider a problem \( P \) characterized by a fixed complexity function \( C_P \). For each input \( x \) in the domain \( D \), there exists a solution \( y \). A implicit pattern \  for problem \( P \), denoted as \( P_{\text{shortcut}} \), is defined as follows:

\begin{itemize}
    \item \( P_{\text{shortcut}} \) is either a subproblem of \( P \) or an independent problem where the domain \( D_{\text{shortcut}} \) is a subset of \( D \) (\ie \( D_{\text{shortcut}} \subseteq D \)).
    \item For any input \( x \) in \( D_{\text{shortcut}} \), the output \( y_{\text{shortcut}} \) of \( P_{\text{shortcut}} \) approximates the output \( y \) of \( P \).
    \item The complexity of solving \( P_{\text{shortcut}} \), \( C_{P_{\text{shortcut}}} \), is significantly less than \( C_P \) (\ie \( C_{P_{\text{shortcut}}} \ll C_P \)).
\end{itemize}
If these conditions are met, then \( P_{\text{shortcut}} \) is considered a shortcut of \( P \). We define its complexity \( C_f \) in terms of the accuracy of a LLM performing on \( f \). Let \( \mathrm{Acc}_f \) represent the accuracy of the LLM on task \( f \), then the complexity \( C_Tf\) can be defined as:
\(
C_T = 1 - \mathrm{Acc}_f
\)
The complexity \( C_f \) ranges from 0 (no complexity, as the task is perfectly solved) to 1 (maximum complexity, as the task is not solved at all).

This definition implies that the higher the LLM's accuracy on a task, the lower the complexity of the task. This measure allows us to quantify task complexity based on the performance capabilities of state-of-the-art language models.

\end{document}